\documentclass[11pt,a4paper]{article}
\usepackage[hyperref]{acl2021}
\usepackage{times}
\usepackage{latexsym}

\usepackage{multirow}
\usepackage{caption}
\usepackage{amssymb}
\usepackage{amsmath}
\usepackage{enumitem}
\usepackage{booktabs}
\usepackage{graphicx}
\usepackage{xcolor}
\usepackage{tabularx}
\usepackage{adjustbox}
\definecolor{royal-blue}{RGB}{65,105,225}
\usepackage[margin=1in]{geometry} 
\newcolumntype{Y}{>{\raggedright\arraybackslash}X} 
\aclfinalcopy 
\def\aclpaperid{3210} 

\newcommand\blfootnote[1]{%
  \begingroup
  \renewcommand\thefootnote{}\footnote{#1}%
  \addtocounter{footnote}{-1}%
  \endgroup
}

\usepackage{microtype}

\title{Towards More Equitable Question Answering Systems: \\ How Much More Data Do You Need?}
\author{Arnab Debnath\textsuperscript{*}, Navid Rajabi\textsuperscript{*}, Fardina Fathmiul Alam\textsuperscript{*}, Antonios Anastasopoulos\\
  Department of Computer Science, George Mason University\\  \texttt{\{adebnath,nrajabi,falam5,antonis\}@gmu.edu}
}
\date{}

\begin{document}
\maketitle

\begin{abstract}

Question answering (QA) in English has been widely explored, but multilingual datasets are relatively new, with several methods attempting to bridge the gap between high- and low-resourced languages using data augmentation through translation and cross-lingual transfer. In this project, we take a step back and study which approaches allow us to take the most advantage of \textit{existing} resources in order to produce QA systems in \textit{many} languages. 
Specifically, we perform extensive analysis to measure the efficacy of few-shot approaches augmented with automatic translations and permutations of context-question-answer pairs. In addition, we make suggestions for future dataset development efforts that make better use of a fixed annotation budget, with a goal of increasing the language coverage of QA datasets and systems.\footnote{Code and data for reproducing our experiments are available here: \url{https://github.com/NavidRajabi/EMQA}.}\blfootnote{\textsuperscript{*}Equal contribution.}

\end{abstract}

\section{Introduction}
\label{section:Introduction}
Automatic question answering (QA) systems are showing increasing promise that they can fulfil the information needs of everyday users, via information seeking interactions with virtual assistants. The research community, having realized the obvious needs and potential positive impact, has produced several datasets on information seeking QA.
The effort initially focused solely on English, with datasets like  WikiQA~\cite{yang2015wikiqa}, MS MARCO~\cite{nguyen2016ms},  SQuAD~\cite{rajpurkar2016squad}, QuAC~\cite{choi2018quac}, CoQA~\cite{reddy2019coqa}, and Natural Questions (NQ)~\cite{kwiatkowski2019natural}, among others. 
More recently, heading calls for linguistic and typological diversity in natural language processing research~\cite{joshi2020state}, larger efforts have produced datasets in multiple languages, such as TyDi QA~\cite{clark2020tydi}, XQuAD~\cite{artetxe2020crosslingual}, or MLQA~\cite{lewis2019mlqa}.

Despite these efforts, the linguistic and typological coverage of question answering datasets is far behind the world's diversity. For example, while TyDi QA includes 11 languages --less than 0.2\% of the world's approximately 6,500 languages~\cite{hammarstrom2015ethnologue}-- from 9 language families, its typological diversity is 0.41, evaluated in a [0,1] range with the measure defined by~\citet{ponti2020xcopa}; MLQA provides data in 7 languages from 4 families, for a typological diversity of 0.32. The total population coverage of TyDi QA, based on population estimates from Glottolog~\cite{nordhoff2012glottolog}, is less than 20\% of the world's population (the TyDiQA languages total around 1.45 billion speakers).


Obviously, the ideal solution to this issue would be to collect enough data in every language. Unfortunately, this ideal seems unattainable at the moment. In this work, we perform extensive analysis to investigate the next-best solution: using the existing resources, large multilingual pre-trained models, data augmentation, and cross-lingual learning to improve performance with just a few or no training examples.
Specifically:
\begin{itemize}
    \item we study how much worse a multilingual few-shot training setting would perform compared to training on large training datasets,
    \item we show how data augmentation through translation can reduce the performance gap for few-shot setting, and
    \item we study the effect of different fixed-budget allocation for training data creation across languages, making suggestions for future dataset creators.
\end{itemize}

\section{Problem Description and Settings}
\label{section:problemdescription}
We focus on the task of simplified minimal answer span selection over a \textbf{gold passage}: The inputs to the model include the full text of an article (the \textit{passage} or \textit{context}) and the text of a question (\textit{query}). The goal is to return the start and end byte indices of the minimal span that completely answers the question. 

Our models follow the current state-of-the-art in extractive question answering, relying on large multilingually pre-trained language models (in our case, multilingual BERT~\cite{devlin2019bert}) and the task-tuning strategy of~\citet{alberti2019bert}, which outperforms approaches like DocumentQA~\cite{clark2018simple} or decomposable attention~\cite{parikh2016decomposable}. In all cases, we treat the official TyDi QA development set as our test set, since the official test set is not public.\footnote{This follows the guidelines to perform analyses over the development set to ensure the integrity of the leaderboard.} We provide concrete details (model cards, hyperparmeters, etc) on our model and training/finetuning regime in Appendix~\ref{app:settings}.

To simulate the scenario of data-scarce adaptation of such a model to unseen languages, we will treat the TyDi QA languages as our test, unseen ones. We will assume that we have access to (a) other QA datasets in more resource-rich languages (in particular, the SQuAD dataset which provides training data in English), and (b) translation models between the languages of existing datasets (again, English) and our target ``unseen" languages.

In the experiments sections, we first focus on few- and zero-shot experiments (\S\ref{sec:fewshot}) and then study the effects of language selection and budget-restricted decisions on training data creation (\S\ref{sec:budget}).

\paragraph{Evaluation} We report F1 score on the test set of each language, as well as a macro-average excluding English (\textbf{$\text{avg}_{\mathcal{L}}$}). In addition, to measure the expected impact on actual systems' users, we follow~\citet{faisal-etal-21-sdqa} in computing a population-weighted macro-average (\textbf{$\text{avg}_{\text{pop}}$}) based on language community populations provided by Ethnologue~\cite{eberhard22simons}. 


\begin{table*}[t]
    \centering
    \begin{tabular}{@{}l@{}|cccccccc|c|c@{\ }}
    \toprule
        & \multicolumn{8}{c|}{\textbf{Results (F1-score)}} &  \textbf{$\text{avg}_{\mathcal{L}}$} & \textbf{$\text{avg}_{\text{pop}}$}\\
    \textbf{Model} & eng & ara & ben & fin & ind & swa & rus & tel & \multicolumn{2}{c}{(without eng)}\\    
    \midrule
    \multicolumn{9}{l}{\small \textbf{Baseline: SQuAD zero-shot}} \\
    (reproduction) & 74.2 & 59.0 & 57.3 & 55.7 & 63.2 & 60.3 & 65.6 & 44.6 & 58.0$\pm$6.3  & 59.3\\
    \midrule 
    \small{Monolingual Few-Shot (+50)} \ \ & 73.9 & 64.9 & 66.4 & 70.9 & 73.3 & 70.1 & 66.3 & 62.5 & 67.8$\pm$3.5 & 67.1\\
    \midrule
    \multicolumn{8}{l}{\small \textbf{Multilingual Few-Shot}} \\
    \hspace{.03cm} (+10/lang, 90 total) & 73.7 & 64.6 & 62.9 & 66.5 & 67.0 & 63.1 & 65.9 & 59.6 & 64.2$\pm$2.4 & 64.4\\
    \hspace{.03cm} (+50/lang, 450 total) & 73.4 & 69.2 & 65.8 & 69.0 & 73.4 & 68.8 & 67.2 & 66.2 & 68.5$\pm$2.4 & 68.6\\
    \hspace{.03cm}  (+100/lang, 900 total) & 74.2 & 72.5 & 70.9 & 71.9 & 75.5 & 72.3 & 69.3 & 69.3 & 71.7$\pm$2.0 & 71.9\\
    \hspace{.03cm}  (+500/lang, 4500 total)\ \ & 76.1 & 76.3 & 74.5 & 78.2 & 81.4 & 79.2 & 73.3 & 73.7 & 76.7$\pm$2.8 & 76.2\\
    \midrule
    \multicolumn{8}{l}{\small \textbf{Data Augmentation + Multilingual Few-Shot}} \\
    +tSQuAD  & 74.9 & 65.4 & 58.4 & 66.7 & 65.2 & 69.4 & 60.2 & 44.7 & 61.4$\pm$7.7 & 61.2 \\
    +mSQuAD & 75.1 & 65.6 & 68.6 & 71.7 & 70.3 &  66.2 & 75.5 & 49.4 & 66.7$\pm$7.7 & 67.6\\
    +mSQuAD  +500/lang & 77.6 & 78.7 & 75.0 & 78.5 & 83.5  & 82.5 & 73.2  & 75.3 & 78.1$\pm$3.6 & 77.6 \\
    +tSQuAD +500/lang& 77.9 & 78.8 & 80.0 & 79.5 & 82.8 & 83.6 & 72.5 & 73.5 & 78.7$\pm$3.9 & 78.6 \\
    \midrule
    \multicolumn{9}{l}{\small \textbf{Skyline: Full training on TyDi QA train}} \\
    (reproduction) & \textit{77.5} & \textit{82.4} & \textit{78.9} & \textit{80.1} & \textit{85.4} & \textit{83.8} & \textit{76.5} & \textit{78.3} & \textit{80.8$\pm$3.0} & \textit{80.9} \\
    \bottomrule
    \end{tabular}
    \caption{Data augmentation combined with multilingual few-shot learning can reach about 98\% of the skyline accuracy using only 10 times less training data on the test languages beyond English.}
    \label{tab:fewshot}
    \vspace{-1em}
\end{table*}

\section{Is Few-Shot a Viable Solution?}
\label{sec:fewshot}

We first set out to explore the effect of the amount of available data on downstream performance. Starting with baselines relying solely on English-only SQuAD, we implement a few-shot setting for fine-tuning on the target languages of TyDi QA.\footnote{We do not report results on Korean, due to a late-discovered issue: we found that parts of the Korean data use a Unicode normalization scheme different than what is expected by mBERT's vocabulary. We suspect this is responsible for our Korean results being consistently around 50\% worse than previously published results.} To our knowledge, this is the first study of its type on the TyDi QA benchmark.

The straightforward baseline simply provides zero-shot results on TyDi QA after training only on English. Table~\ref{tab:fewshot} provides our (improved) reproduction of the baseline experiments of~\citet{clark2020tydi}. 
The skyline results (bottom of Table~\ref{tab:fewshot}) reflect the presumably best possible results under our current modeling approach, which trains jointly on all languages using all available TyDi QA training data. We note that for most languages the gap between the baseline and the skyline is more than 20 percentage points, with the exception of English where --unsurprisingly-- there is a difference of only $3.3$ percentage points. The performance gap is smallest for Russian (rus) at 10.9 percentage points, and largest for Telugu (tel) at 34 points.

We first study a \textit{monolingual} few-shot setting. That is, we fine-tune the model trained on the English SQuAD dataset, with only a small amount of data (10, 20, or 50 training instances) in the test language. Due to space limitations, we only present results with 50 examples per language in Table~\ref{tab:fewshot}, but the full experiments are available in Appendix~\ref{app:fewshot}. We observe that even just 50 additional training instances are enough for significant improvements, which are consistent across all languages. For example, the improvement in Finnish (fin) exceeds 15 percentage points and covers about more than 60\% 
of the performance gap between the baseline and the skyline.  

We now turn to a \textit{multilingual} few-shot setting. Exactly as before, we assume a scenario where we only have access to a small amount of data in each language, but now we fine-tune using that small amount of data in all languages. For example, 10 training instances in each language result in training with 90 training examples over the 9 test languages. A sample of our experimental results are presented in Table~\ref{tab:fewshot} under ``multilingual few-shot," with complete results in Appendix~\ref{app:fewshot}.

Simply adding 50 instances from each language we obtain an  F1 score  of  67.9 over the zero-shot baseline, an improvement of almost 7 percentage points which reduces the zero-full gap by 43.4\%. We note that the total 450 training instances represent less than 1\% of the full TyDi QA training set! Doubling that amount of data to 100 examples per language further increases downstream performance to an average overall F1 score of 71.7. Going further to the point of adding 500 training instances per language (for a total of 4500 examples) leads to even larger improvements for an average F1 score of 76.7. That is, using less than 10\% of the available training data  
we can reduce the average F1 score performance gap by more than 82\%. 
For a few languages the gap reduction is even more notable, e.g., more than 92\% for Finnish. 

\paragraph{Data Augmentation through Translation}
Generating translations of English dataset to train systems in other languages has a long history and has been successful in the QA context as well~\cite[\textit{inter alia}]{yarowsky2001inducing,xue2020mt5}. We follow the same approach, translating all SQuAD paragraphs, questions, and answers to all TyDi QA languages using Google Translate.\footnote{We release the data to facilitate the reproduction of our experiments.} For each language, we keep between 20-50\% of the question-answer pairs where the translated answer has an exact match in the translated paragraph, which becomes the target span.\footnote{This approach could be enhanced using word/phrase alignment techniques, which we leave for future work.} 
Details of the resulting dataset (which we refer to as tSQuAD) are in 
Table~\ref{tab:tranlated_dataset_information} in 
Appendix~\ref{app:augmentation}.
A second approach translates the question of a training instance into one language, but keeps the answer and context into the original language. The result is a modified training set (which we name mSQuAD) that requires better cross-lingual modeling, as the question and contexts are in different languages. 

Both approaches improve over the zero-shot baseline with F1 score of 61.4 (+3) and 66.7 (+8). Notably, though, they are not as effective as few-shot training even with just 50 instances per languages. This further strengthens the discussion of~\citet{clark2020tydi} on the qualitative differences between the SQuAD and TyDi QA dataset. Nevertheless, combining tSQuAD (or mSQuAD) with a few examples from the TyDi QA dataset leads to our best-performing methods. In particular, augmentation through translation leads to an 1-2 percentage point improvements over the multilingual few-shot approach (cf. 76.7 to 78.1/78.7 F1 score in Table~\ref{tab:fewshot}; full results in Appendix~\ref{app:fewshot}). Now, using only 500 new training examples per language we are \textit{almost} (98\%) at similar performance levels as the skyline.

\begin{table*}[t]
    \centering
    \small
    \begin{tabular}{@{}c@{ \ }c@{ \ }c@{ \ }c@{ \ }c@{ \ }c@{ \ }c@{ \ }c@{ }|@{ }c@{ }|@{ }c@{ }|@{ }c@{ }c@{}}
    \toprule
    \multicolumn{7}{c}{\textbf{\hspace{1cm} Results (F1-score)}} & &  \textbf{Overall} & \textbf{$\Delta_l$} & \multicolumn{2}{c}{avg}\\
    eng & ara & ben & fin & ind & swa & rus & tel & \small{(w/o eng)} & \footnotesize{(max-min)} & \footnotesize{seen} & \footnotesize{unseen}\\
    \midrule
    \multicolumn{9}{l}{\small \textbf{Baseline: no budget for additional data (zero-shot except for eng)}} \\
    74.2 & 59.0 & 57.3 & 55.7 & 63.2 & 60.3 & 65.6 & 44.6 & 58.0{\small{$\pm$6.3}} & 29.6 & 74.2 & 58.0  \\
    \midrule 
    \multicolumn{9}{l}{\small \textbf{Monolingual budget allocation (max 4500 per language; 7 experiments)}}\\
    76.0\small{$\pm$1.8} & 74.0\small{$\pm$3.9} & 69.1\small{$\pm$5.0} & 75.8\small{$\pm$2.7} & 78.4\small{$\pm$4.1} & 71.7\small{$\pm$4.1} & 75.7\small{$\pm$6.3} & 61.3\small{$\pm$12.3} & 72.3\small{$\pm$5.3} & 17.1 & 77.1	& 71.3\\
    \midrule
    \multicolumn{9}{l}{\small \textbf{Tri-lingual budget allocation (1500 per language; 7 random language selection experiments)}} \\
     76.7\small{$\pm$1.2} & 77.2\small{$\pm$2.8} & 68.6\small{$\pm$4.8} & 77.9\small{$\pm$1.6} & 80.9\small{$\pm$3.3} & 81.5\small{$\pm$3.3} & 72.7\small{$\pm$2.3} & 62.9\small{$\pm$13.3} & 74.5\small{$\pm$6.3} & 18.6 & 78.9 & 68.5\\
    \midrule
    \multicolumn{9}{l}{\small \textbf{Uniform budget allocation (500 per language)}} \\
    77.9 & 78.8 & 80.0 & 79.5 & 82.8 & 83.6 & 72.5 & 73.5 & 78.7\small{$\pm$3.9} & 11.1 & 78.6 & -- \\
    \midrule
    \multicolumn{10}{l}{\small \textbf{\textit{Ideal Few-Shot (4500 in each language; in-language results})}} \\
    \textit{78.4} & \textit{81.8} & \textit{77.7} & \textit{79.7} & \textit{83.9} & \textit{84.0} & \textit{75.7} & \textit{78.2} & \textit{79.9\small{$\pm$3.0}} & \textit{8.3} & \textit{79.9} & -\\
    \bottomrule
    \end{tabular}
    \caption{A more egalitarian budget allocation leads to better \textit{and} more equitable performance across languages ({\small avg$\pm$std: higher average, lower std.\ deviation}) reducing the gap ($\Delta_l$) between best and worst performing languages.}
    \label{tab:budget}
    \vspace{-1em}
\end{table*}

\section{How to Spend the Annotation Budget?}
\label{sec:budget}

In the previous section we show that the combination of data augmentation techniques with a few new annotations 
can reach almost 98\% of the performance one would obtain by training on 10x more data. In this section we explore how one should allocate a fixed annotation budget, in order to achieve not only higher average but also more \textit{equitable} performance across languages.

Keeping our budget fixed to 4500 instances, we study 3 scenarios. The first is \textbf{monolingual} allocation, where the whole budget is consumed by collecting training examples on a single language. 
We repeat the study over all 8 languages of our test set, randomly sampling training instances from the TyDi QA training set. Second, we study a \textbf{tri-lingual} budget allocation scheme, where we equally split the budget across 3 languages for 1500 training instances per language. We repeat this experiment 7 times, each time randomly selecting 3 languages. Last, the third and more \textbf{egalitarian} scenario splits the budget equally across all 8 languages, matching our previously analyzed few-shot scenario where we only have 500 additional training examples per language.
In all experiments, we use our best-performing approach from the previous section, also utilizing tSQuAD for pre-training.

Our findings are summarized in Table~\ref{tab:budget}. For the repeated monolingual and tri-lingual scenarios we report average performance across our experiment repetitions (full results in Appendix~\ref{app:budget}). We can conclusively claim that a uniform budget allocation leads to not only better average performance, but also to more equitable performance. We report two straightforward measures for the equitability of the average accuracy across languages. First, we report the standard deviation of the accuracy across languages; the lower the standard deviation, the more equitable the performance. We also report the difference between the best and the worst performing language for each experiment, as well as the averages for the languages that are seen and unseen during fine-tuning.

Having no budget for additional annotation (essentially, attempting the task in zero-shot fashion) leads to the most inequitable performance. The monolingual scenario typically leads to the highest accuracy when evaluating on the same language as the new training examples (the \textit{ideal} section of Table~\ref{tab:budget}) but the zero-shot performance on all other languages is generally significantly worse, leading to inequity. The tri-lingual scenarios follow similar patterns, with performance close to state-of-the-art for the four languages (three plus English) that have been included in the fine-tuning process, but with the rest of the languages lagging behind: the difference between seen and unseen languages is on average 10.4 points. In our experiments we randomly sampled (without replacement) three of the seven languages, but one could potentially use heuristics or a meta-model like that of~\citet{xia-etal-2020-predicting} to find or suggest the best subset of candidate languages for transfer learning; we leave such an investigation for future work.

Encouragingly, the uniform budget allocation scenario leads to higher average performance, while also reducing the gap between worst and best performing languages from around 30 percentage points to less than 12 points (60\% reduction). Note that a 8x larger budget (\textit{ideal} scenario) with 4500 instances per language would further improve downstream accuracy and equitability. Note that in this case where some resources are available, simple multilingual fine-tuning might not be the best approach for some languages, e.g. compared to monolingual fine-tuning or meta-learning approaches~\cite[\textit{inter alia}]{wang-etal-2020-negative,muller-etal-2021-unseen}. We leave an investigation of such settings for future work. 

\section{Discussion}
We show that data augmentation through translation along with few-shot fine-tuning on new languages with a uniform budget allocation leads to a performance close to 98\% of an approach using 10x more data, while producing more equitable models than other budget-constrained alternatives.

The implications of our findings become clear with a counter-factual exploration. The Gold Passage portion of the TyDi QA dataset includes around 87,000 annotated examples (50k for training across 9 languages and about 37k development and test samples). Consider the scenario where, given this annotation budget, we maintain the same evaluation standards collecting 4k development and test examples per language, but we only collect 500 training examples per language. In that case, we could have created a much more diverse resource that would include at least 19 languages! Now consider the expectation of the downstream accuracy in our counterfactual scenario: uniform budget allocation on 19 languages would lead to an average accuracy (F1 score) of around 78\% (similar to our experiments). Instead, under the (currently factual) scenario where we only have training data for 9 languages, the average accuracy for these 9 languages is around 80\%, but the zero-shot expected average on the other 10 languages is 10 points worse -- in that case, the overall average accuracy would be around 74\%, 4 points lower than that of the egalitarian allocation scenario.
Hence, as long as the ideal scenario of collecting a lot of data for a lot of languages remains infeasible, we suggest that the community puts an additional focus on the linguistic diversity of our evaluation sets and use other techniques to address the lack of training data.

\section*{Acknowledgements}
This work is supported by NSF Award 2040926. The authors also want to thank Fahim Faisal for helpful discussions on setting up the experiments. Most experiments were run on ARGO,\footnote{\url{http://orc.gmu.edu}} a research computing cluster provided by the Office of Research Computing at George Mason University, VA, and a few experiments were run on Amazon Web Services instances donated through the AWS Educate program.

\bibliography{acl2020}
\bibliographystyle{acl_natbib}
\appendix
\section{Experimental Settings}
\label{app:settings}

For the experiments, we've used ``\texttt{bert-multi-lingual-base-uncased}" (mBERT)  \cite{huggingface:mBERT}  as mentioned as the main baseline on TyDi QA paper \cite{clark2020tydi}. It is a pre-trained model on the top 102 languages with the largest Wikipedia using a masked language modeling (MLM) objective \cite{devlin2019bert}. 
From preliminary experiments, we realized that the optimum trade-off between the highest F1 score and the least computational cost is achieved by training for 3 epochs, using batch size of 24, and learning rate of 3e-5. Therefore, we applied these hyperparameter settings for our experiments.
The main script we used was a module under the Huggingface library \cite{Wolf2019HuggingFacesTS} (called run\_squad
), which is being used widely for fine-tuning transformers for multi-lingual question answering datasets.

\section{SQuAD Translation Details}
\label{app:augmentation}
We augmented the English SQuAD with \textit{translated SQuAD} (tSQuAD) instances for each language.  Here, the contexts, questions and answers from SQuAD instances are translated to the target languages using Google Translate (with the \texttt{google-trans-new} API) and only the instances where an exact match of translated answer is found in the translated context, are kept for augmentation. The total number of instances per language, we ended up with after translation is listed in Table~\ref{tab:tranlated_dataset_information}.

\begin{table*}[t]
    \centering
    \small
    \begin{tabular}{l|c|cccccccc}
    \toprule
     & SQuAD & tAr & tBn & tFin & tInd & tKo & tRus & tSwa & tTel \\    
    \midrule
    no of paragraphs & 18.9 & 16.6 & 13.5 & 12.4 & 16.2 & 11.2 & 11.6 & 15.3 & 16.6 \\
    \midrule 
    no of QAs & 87.6 & 39.1 & 24.1 & 21.4 & 36.1 & 18.1 & 19.2 & 31.2 & 39.7  \\
    \bottomrule
    \end{tabular}
    \caption{Number (in 1000s) of paragraphs and QA pairs present in the original SQuAD and translated SQuAD}
    \label{tab:tranlated_dataset_information}
\end{table*}

\section{Complete Few-Shot Experiments}
\label{app:fewshot}

Provided in Table~\ref{tab:fewshot2}.
\begin{table*}[t]
    \centering
    \small
    \begin{tabular}{l|cccccccc|c}
    \toprule
        & \multicolumn{8}{c}{\textbf{Results (F1-score)}} &  \textbf{Overall}\\
    \textbf{Model} & eng & ara & ben & fin & ind & swa & rus & tel & (without eng)\\    
    \midrule
    \multicolumn{9}{l}{\small \textbf{Baseline: SQuAD zero-shot}} \\
    \cite{clark2020tydi} & 73.4 &  60.3 & 57.3 & 56.2 & 60.8 & 52.9 & 64.4 & 49.3 & 57.3$\pm$4.7 \\
    (ours) & 74.2 & 59.0 & 57.3 & 55.7 & 63.2 & 60.3 & 65.6 & 44.6 & 58.0$\pm$6.3 \\
    \midrule 
    Monolingual Few-Shot (+10) & 73.7 & 64.7 & 62.8 & 68.2 & 69.3 & 59.9 & 65.6 & 50.7 & 63.0$\pm$5.8 \\
    Monolingual Few-Shot (+20) & 74.7 & 63.5 & 60.5 & 66.6 & 72.1 & 63.9 & 66.8 & 63.0 & 65.2$\pm$3.4 \\
    Monolingual Few-Shot (+50) & 73.9 & 64.9 & 66.4 & 70.9 & 73.3 & 70.1 & 66.3 & 62.5 & 67.8$\pm$3.5 \\
    \midrule
    \multicolumn{8}{l}{\small \textbf{Multilingual Few-Shot}} \\
    \hspace{.5cm} (+10/lang, 90 total) & 73.7 & 64.6 & 62.9 & 66.5 & 67.0 & 63.1 & 65.9 & 59.6 & 64.2$\pm$2.4\\
    \hspace{.5cm} (+20/lang, 180 total) & 73.9 & 65.9 & 66.8 & 69.0 & 72.5 & 64.2 & 66.9 & 63.7 & 67.0$\pm$2.8 \\
    \hspace{.5cm} (+50/lang, 450 total) & 73.4 & 69.2 & 65.8 & 69.0 & 73.4 & 68.8 & 67.2 & 66.2 & 68.5$\pm$2.4\\
    \hspace{.5cm} (+100/lang, 900 total) & 74.2 & 72.5 & 70.9 & 71.9 & 75.5 & 72.3 & 69.3 & 69.3 & 71.7$\pm$2.0\\
    \hspace{.5cm} (+200/lang, 1800 total) & 73.9 & 74.8 & 70.5 & 74.1 & 77.7 & 76.4 & 69.8 & 70.0 & 73.3$\pm$3.0\\
    \hspace{.5cm} (+500/lang, 4500 total)& 76.1 & 76.3 & 74.5 & 78.2 & 81.4 & 79.2 & 73.3 & 73.7 & 76.7$\pm$2.8\\
    \midrule
    
    \multicolumn{8}{l}{\small \textbf{Data Augmentation + Multilingual Few-Shot}} \\
   
    +tSQuAD(50/lang)& 73.8 & 64.0 & 62.4 & 68.4 & 69.7 & 59.7  & 66.8 & 48.1 & 62.7$\pm$6.8\\
    +tSQuAD(100/lang) & 72.4 & 62.2 & 66.6 & 68.4 & 68.6 & 64.9 & 67.1 & 47.5 & 63.6$\pm$6.9\\
    +tSQuAD(200/lang)& 74.4 & 62.7 & 64.2 & 68.8 & 70.7 &  66.1 & 66.2 & 48.3 & 63.9$\pm$6.8\\
    +tSQuAD(500/lang)& 73.7 & 63.2 & 69.5 & 67.9 & 70.9 & 69.8 & 66.7 &  49.1 & 65.3$\pm$7.0\\
    
    +tSQuAD(all)  & 74.9 & 65.4 & 58.4 & 66.7 & 65.2 & 69.4 & 60.2 & 44.7 & 61.4$\pm$7.7\\
     +mSQuAD  +500/lang & 77.6 & 78.7 & 75.0 & 78.5 & 83.5  & 82.5 & 73.2  & 75.3 & 78.1$\pm$3.6   \\
    +tSQuAD +500/lang (mBERT)& 77.9 & 78.8 & 80.0 & 79.5 & 82.8 & 83.6 & 72.5 & 73.5 & 78.7$\pm$3.9\\
    +tSQuAD +500/lang (XLM-R)$^*$& 73.2 & 72.8 & 78.3 & 78.5 & 84.7 & 80.3 & 75.0 & 78.1 & 78.2$\pm$3.5 \\
    \midrule
    \multicolumn{9}{l}{\small \textbf{Skyline: Full training on TyDi QA train}} \\
    \cite{clark2020tydi} & 76.8 & 81.7 & 75.4 & 79.4 & 84.8 & 81.9 & 76.2 & 83.3 & 80.4$\pm$3.3\\
    (ours) & 77.5 & 82.4 & 78.9 & 80.1 & 85.4 & 83.8 & 76.5 & 78.3 & 80.8$\pm$3.0\\
    \bottomrule
    \end{tabular}
    \caption{Complete few-shot and data augmentation results. $*$: Results with XLM-Roberta-Large~\cite{conneau2019unsupervised} are generally worse than using mBERT so all other experiments use mBERT.}
    \label{tab:fewshot2}
\end{table*}

\section{Mix-and-Match Experiments}
\label{app:mixmatch}

Provided in Table~\ref{tab:mixmatch}.

\begin{table*}[t]
\begin{tabularx}{\textwidth}{@{}lYYYYYY@{}} 
\toprule

&\multicolumn{3}{c}{\bfseries Change language of Question only }
&\multicolumn{3}{c}{\bfseries Change all; Context \& answers the same} \\

\cmidrule(lr){2-4} \cmidrule(l){5-7} 

& Modified Squad 
& Squad + Modified Squad 
& Squad + Modified Squad + 500 instances 
& Modified Squad 
& Squad + Modified Squad 
& Squad + Modified Squad + 500 instances \\
\midrule
English & 66.59 & 75.06 & 77.56 & 65.40 & 73.49 & 78.21 \\
Arabic & 62.17 & 65.62 & 78.70 & 60.51 & 65.98 & 77.96  \\
Bengali & 67.33 & 68.55 & 75.00 & 58.60 & 62.44 & 76.16  \\
Finnish & 67.42& 71.67 & 78.55 & 62.98 & 67.58 & 79.51   \\
Indonesian & 66.45 & 70.33 & 83.46 & 61.89 & 66.44 & 84.10 \\
Kiswahili & 70.32 & 75.48 & 82.51 & 62.66 & 68.55 & 80.01  \\
Russian & 64.71 & 66.16 & 73.16 & 61.01 &  65.64 & 73.28  \\
Telugu & 48.32 & 49.36 & 75.28 & 43.62 & 51.81 & 74.95  \\
\midrule
Avg & 63.82 & 66.74 & \textbf{78.09} & 58.76 & 64.07 & \textbf{78.00} \\
SD & 6.74 & 7.75 & 3.60 & 6.33 & 5.31 & 3.35 \\
\bottomrule
\end{tabularx}
\caption{Mix-and-Match scheme detailed results.} \label{tab:mixmatch}
\end{table*}

\section{Budget Allocation Experiments}
\label{app:budget}

The complete results for our experiments are presented in Table~\ref{tab:budget2}.

\begin{table*}[t]
    \centering
    \small
\adjustbox{max width=\textwidth}{%
    \begin{tabular}{@{}c@{}c@{ \ }c@{ \ }c@{ \ }c@{ \ }c@{ \ }c@{ \ }c@{ \ }c@{ }|@{ }c|c@{ }c@{}}
    \toprule
    &\multicolumn{8}{c}{\textbf{Results (F1-score)}} & \textbf{Overall} & \multicolumn{2}{c}{\textbf{Avg}} \\
    & eng & ara & ben & fin & ind & swa & rus & tel & \small{(w/o eng)} & seen & unseen \\  
    \midrule
    \multicolumn{10}{l}{\small \textbf{Baseline: no budget for additional data (zero-shot excelt in eng)}} \\
    & 74.2 & 59.0 & 57.3 & 55.7 & 63.2 & 60.3 & 65.6 & 44.6 & 60.0{\small{$\pm$8.5}} & 74.2 & 58.0 \\
    \midrule 
    \multicolumn{10}{l}{\small \textbf{Monolingual budget allocation (max 4500 per language; 7 experiments)}}\\
    Arabic & 78.4 & 81.8 & 62.0 & 77.6 & 79.2 & 72.8 & 68.0 & 50.5 & 70.2$\pm$10.3 & 80.1 & 68.4 \\
    Bengali & 74.4 & 66.3 & 77.7 & 71.6 & 72.8 & 78.1 & 66.5 & 52.0 & 69.3$\pm$8.3 & 76.1 & 67.9  \\
    Finnish & 77.9 & 75.5 & 72.6 & 79.7 & 81.0 & 70.6 & 78.5 & 52.2 & 72.9$\pm$9.1 & 78.8 & 71.7 \\
    Indonesian & 76.8 & 76.7 & 67.4 & 77.0 & 83.9 & 70.2 & 77.3 & 52.2 & 72.1$\pm$9.5 & 80.4 & 70.1\\
    Kiswahili & 76.4 & 72.5 & 67.1 & 75.0 & 77.4 & 66.4 & 84.0 & 75.0 & 73.9$\pm$5.6 & 71.4 & 75.2 \\
    Russian & 75.2 & 74.5 & 66.7 & 76.3 & 81.0 & 75.7 & 78.8 & 69.4 & 74.6$\pm$4.7 & 77.0 & 73.9 \\
    Telugu & 73.4 & 70.6 & 70.2 & 73.6 & 73.7 & 68.1 & 77.1 & 78.2 & 73.1$\pm$3.4 & 75.8 & 72.2 \\
    & 76.0\small{$\pm$1.8} & 74.0\small{$\pm$3.9} & 69.1\small{$\pm$5.0} & 75.8\small{$\pm$2.7} & 78.4\small{$\pm$4.1} & 71.7\small{$\pm$4.1} & 75.7\small{$\pm$6.3} & 61.3\small{$\pm$12.3} & 72.3\small{$\pm$5.3} & 77.1	& 71.3\\
    \midrule
    \multicolumn{10}{l}{\small \textbf{Tri-lingual budget allocation (1500 per language; 7 random language selection experiments)}} \\
    ben-rus-tel & 75.8 & 72.2 & 79.0 & 75.6 & 74.8 & 77.1 & 74.5 & 76.8 & 75.7$\pm$2.0 & 76.5 &	74.9\\
    tel-ind-swa & 76.1 & 75.7 & 65.5 & 76.7 & 83.2 & 84.7 & 71.2 & 77.2 & 76.3$\pm$6.1 & 80.3 & 72.3\\
    fin-rus-swa & 78.5 & 76.4  & 66.3  & 79.6 & 80.3 & 84.8 & 74.9 & 53.4 & 73.7$\pm$9.8 & 79.5	& 69.1  \\
    ara-rus-tel & 75.7 & 79.3 & 66.8 & 78.0 & 79.2 & 79.9 & 74.3 & 77.0 & 76.4$\pm$4.3 & 76.6 & 60.8 \\
    ara-rus-fin & 76.5 & 80.5 & 68.9 &	79.2 & 80.6 & 77.5 & 74.3 & 53.6 & 73.5$\pm$9.0 & 77.6 & 70.2 \\
    swa-ind-fin & 76.1 & 77.2 & 68.5 & 79.7 & 84.2 & 83.0 & 71.2 & 51.5 & 73.6$\pm$10.5 & 80.8	& 67.1  \\
    ara-ind-swa & 78.3 & 79.5 & 65.4 & 76.8 & 83.9 & 83.5 & 68.9 & 50.6 & 72.7$\pm$11.1 & 81.3 & 65.4\\
    & 76.7\small{$\pm$1.2} & 77.2\small{$\pm$2.8} & 68.6\small{$\pm$4.8} & 77.9\small{$\pm$1.6} & 80.9\small{$\pm$3.3} & 81.5\small{$\pm$3.3} & 72.7\small{$\pm$2.3} & 62.9\small{$\pm$13.3} & 74.5\small{$\pm$6.3}  & 78.9 & 68.5\\
    \midrule
    \multicolumn{9}{l}{\small \textbf{Uniform budget allocation (500 per language)}} \\
    & 77.9 & 78.8 & 80.0 & 79.5 & 82.8 & 83.6 & 72.5 & 73.5 & 78.7\small{$\pm$3.9} & 78.6 & - \\
    \bottomrule
    \end{tabular}}
    \caption{Complete budget allocation experiments.}
    \label{tab:budget2}
\end{table*}

\end{document}